\documentclass{article}
\usepackage{spconf,amsmath,graphicx}
\usepackage{soul,xcolor}
\usepackage{enumerate}
\usepackage{adjustbox}
\usepackage{multirow}
\usepackage{tabularx}
\usepackage{hyperref}

\title{Speech Tasks Relevant to Sleepiness Determined with Deep Transfer Learning}

\name{Bang Tran$^1$, Youxiang Zhu$^1$, Xiaohui Liang$^1$, James W. Schwoebel$^2$, Lindsay A. Warrenburg$^2$\thanks{This research is funded by the US National Institutes of Health National Institute on Aging, under grant No. R01AG067416.}}
\address{$^1$Department of Computer Science, University of Massachusetts Boston, MA, USA\\ $^2$Sonde Health Inc.}

\begin{document}
\maketitle
\begin{abstract}

Excessive sleepiness in attention-critical contexts can lead to adverse events, such as car crashes. Detecting and monitoring sleepiness can help prevent these adverse events from happening. In this paper, we use the Voiceome dataset to extract speech from 1,828 participants to develop a deep transfer learning model using Hidden-Unit BERT (HuBERT) speech representations to detect sleepiness from individuals. Speech is an under-utilized source of data in sleep detection, but as speech collection is easy, cost-effective, and non-invasive, it provides a promising resource for sleepiness detection. Two complementary techniques were conducted in order to seek converging evidence regarding the importance of individual speech tasks. Our first technique, \textit{masking}, evaluated task importance by combining all speech tasks, masking selected responses in the speech, and observing systematic changes in model accuracy. Our second technique, \textit{separate training}, compared the accuracy of multiple models, each of which used the same architecture, but was trained on a different subset of speech tasks. Our evaluation shows that the best-performing model utilizes the \textit{memory recall} task and \textit{categorical naming} task from the Boston Naming Test, which achieved an accuracy of 80.07\% (F1-score of 0.85) and 81.13\% (F1-score of 0.89), respectively.

\end{abstract}


\section{Introduction}
\label{sec:intro}
Sleepiness results in human cognitive and behavioral changes that may cause a variety of negative outcomes, including automobile crashes, poor work performance, accidents at work, and other long-term physical and mental health consequences~\cite{nollet2020sleep,RR1791VH}. As one example, the U.S. National Highway Traffic Safety Administration reported that in 2019 alone, 697 deaths were due to sleepiness-related automobile crashes~\cite{NHTSA:report2019}. The following year, a poll conducted by the National Sleep Foundation showed that nearly half of Americans feel sleepy on an average of three days a week~\cite{sleepfoundation:poll2020}. Given these statistics, it is clear that excessive sleepiness negatively impacts safety and threatens our working and living environment.

By detecting sleepiness, it is possible to alert people to unsafe conditions and allow them to evaluate their own capacity to engage in dangerous behavior, such as driving. Analyzing speech for signs consistent with sleepiness is an ideal solution, as speech collection is non-invasive, low-cost, scalable, and can be performed quickly and easily~\cite{schuller2019interspeech}. One practical barrier with speech-based sleepiness detection has been the limited availability of human speech datasets. The ComParE 2019 challenge, for example, utilized speech recordings from 915 German speakers with two types of speech elicitations (reading and spontaneous speech). While this challenge resulted in significant contributions in sleepiness detection, the models and results were limited due to the low number of speech tasks and participants~\cite{mielke2014clinical}. 

The current study used speech samples from the Voiceome dataset, which includes speech data from 6,650 participants. In the Voiceome dataset, each participant responded to 12 types of speech tasks (e.g., picture description, category naming, memory recall) for a total of 48 speech utterances. The participants also answered questions about their mental and physical health, including sleepiness, depression, anxiety, and smoking status.

The primary goal of the current paper was to infer a person's self-reported sleepiness from their speech. Secondary goals included (1) evaluating the importance of each of the 12 speech tasks in the detection process using a \textit{masking} technique and (2) confirming and evaluating the potential of using a single speech task for sleepiness detection using a \textit{separate training} technique. 

\section{Related works}
\label{sec:related_works}

Detecting sleepiness has been addressed in past research by investigating acoustic speech factors. Previous methods have used a large number of general-purpose low-level descriptors (LLDs) such as short-term spectrum, short-term energy, and other voice-related features. Schuller et al. summarized these previous methods from the Interspeech 2011 Speaker State Challenge on sleepiness estimation~\cite{schuller2014medium}. Histogram representations of clustered LLDs, known as bag-of-audio-words, and melspectrogram feature representations were studied in the Interspeech 2019 ComParE challenge~\cite{schuller2019interspeech}. For example, Yeh et al. presented a system that uses eGeMAPS features as the input of a Bidirectional Long Short-Term Memory network with attention to estimate sleepiness levels from speech. Gosztolya et al. created utterance-level Fisher vectors by training a Gaussian Mixture Model on frame-level MFCC features and used the vectors for sleepiness classification with Support Vector Machine~\cite{gosztolya2019using}. To address acoustic-phonetic changes in sleepy speech, Fritsch et al. investigated the differences in speech production from a phonetic perspective by inheriting the knowledge of a pre-trained Convolutional Neural Network model to extract articulatory features from the speech data~\cite{fritsch2020estimating}.  Egas-López et al. addressed the same problem by adopting a pre-trained x-vector model to estimate sleepiness level~\cite{egas2021deep}. 

In this paper, we employ the Hidden-Unit BERT~\cite{hsu2021hubert} speech representation technique to extract acoustic-phonetic and linguistic features from speech data for sleepiness classification.

\begin{figure*}[!ht]
    \centering
    \centerline{\includegraphics[width=0.9\linewidth]{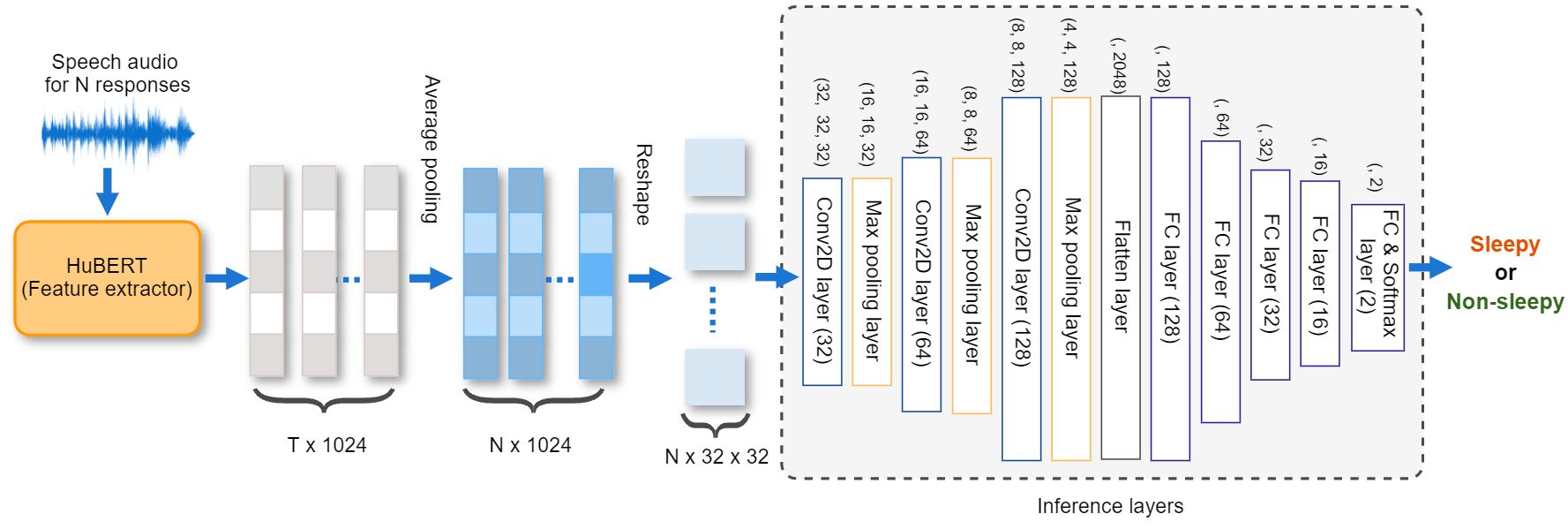}}
    \vspace{-0.2cm}
    \caption{HuBERT-Sleep deep transfer learning model.}
    \label{fig:system_model}
    \vspace{-0.5cm}
\end{figure*}

\section{Materials}
\label{sec:dataset}
\subsection{Voiceome protocol and dataset}
The Voiceome protocol is a high-fidelity, longitudinal, and scalable protocol that can be used in digital settings to advance speech and language biomarkers research~\cite{schwoebel2021longitudinal}. The corresponding Voiceome dataset consists of responses from 6,650 participants who completed the Voiceome protocol. All participants were residents of the United States, were 18 years of age or older, and indicated that they felt comfortable reading and writing in English. Additionally, all participants were required to have access to a device with a microphone and internet connection. The participants were broadly representative of the U.S. population with respect to age, gender, race and ethnicity, and general health behaviors~\cite{schwoebel2021longitudinal}. More details can be found on the Voiceome GitHub\footnote{https://github.com/jim-schwoebel/voiceome}.

The Voiceome protocol consists of twelve speech tasks, which result in a total of 48 speech utterances. The tasks are as follows:

1. \textit{Microphone test}: read the sentence, "The quick brown fox jumps over lazy dog" (10 sec).

2. \textit{Free speech}: talk about a recent happy memory based on experiences from the past month (60 sec).

3. \textit{Picture description}: look at an image and describe everything they see going on in the picture (60 sec).

4. \textit{Category naming}: given a category, such as "animals," speak as many words in that category as they can (60 sec).

5. \textit{Phonemic fluency}: given a letter, such as "F," speak as many words that start with that letter as they can (60 sec).

6. \textit{Phonetically-balanced paragraph reading}: read the Caterpillar passage~\cite{patel2013caterpillar}.

7. \textit{Sustained phonation}: make the vowel sound "/\textit{a}/" (as in hallelujah) for as long as they can (30 sec).

8. \textit{Diadochokinetic task}: repeat \textit{puh-puh-puh} as quickly and accurately as they can (10 sec).

9. \textit{Diadochokinetic task}: repeat \textit{puh-tuk-kuh} as quickly and accurately as they can (10 sec).

10. \textit{Confrontational naming}: look at an image and speak the name of the image within 10 seconds; 25 images total.

11. \textit{Non-word pronunciation}: pronounce the "non-word" (e.g., plive, fwov) shown on the screen within 10 seconds; 10 non-words total.

12. \textit{Memory recall}: listen to a short audio clip (15 sec) and repeat the sentence that they just heard.

\subsection{Sleepiness operationalization}
Voiceome participants completed the Stanford Sleepiness Scale (SSS), which is a single question that asks people to self-report their current state. The SSS is a Likert scale that ranges from \textit{1--Feeling active, vital, alert, or wide awake} through \textit{7--No longer fighting sleep, sleep onset soon; having dream-like thoughts}. The middle of the scale is biased towards sleepiness, with an indicator of \textit{4--Somewhat foggy, let down}. For the purposes of the current study, we operationalized sleepiness as a binary category, where sleepiness ratings of 1-3 indicate non-sleepy states and ratings of 4-7 represent sleepy states. Note that only half of the Voiceome participants included health annotations in the Voiceome dataset; of those participants, 1,828 participants completed every speech task.

\subsection{Speech pre-processing}
The 1,828 participants submitted a total of 186.2 hours of speech responses. Voice samples went through similar pre-processing steps, including being converted to mono recordings at a sampling rate of 16,000 Hz. The Voiceome GitHub page provides all code used for audio pre-processing, feature extraction, and automatic transcription.

\section {System Design}
\label{sec:system_model}
Deep transfer learning has been proven effective across a wide range of domains, outperforming traditional machine learning approaches. We used the HuBERT model recently released by Facebook to featurize speech files to create a fixed-length embedding (1024 dimensions per audio file). We also separately benchmarked this deep learning model across a TPOT-trained classifier~\cite{olson2016tpot} with the OpenSMILE GeMAPSv01a~\cite{eyben2015geneva} aggregate feature embedding (62 dimensions per audio file). These approaches are discussed in the sections that follow.

\vspace{-0.2cm}

\subsection{HuBERT-Sleep Architecture/Baseline model}
\label{subsec:baseline_model}

The HuBERT-Sleep baseline model was developed by incorporating the HuBERT pre-trained model as a fixed-length feature extractor (1024 dimensions), with added inference layers for the downstream task as shown in Figure~\ref{fig:system_model}. A fixed feature extractor was selected to minimize overfitting because the dataset of the downstream task was relatively small~\cite{zhu2021exploring}. The HuBERT embedding was selected because it is the state-of-the-art approach for speech representations~\cite{yang2021superb, lakhotia2021generative, polyak2021speech}.

For each session, a participant has $N$ speech utterances, with a total possibility of 48 responses. The $N$ responses of a session were concatenated into a single speech input. If the speech input has $T$ frames, it would be denoted  by $X = [x_1,\dots, x_T ]$. The fixed-length HuBERT feature extractor outputs 1024 hidden units for each frame $x_i$. The hidden units for all frames of $X$ were denoted as $Z = [z_1, \dots ,z_T ]$, where $z_i$ is a 1024-dimension vector. An average pooling layer was applied to the speech frames from the same response in order to reduce the dimensionality of $Z$ from $T \times 1024$ to $N \times 1024$. The $N$ vectors of dimension 1024 were mapped onto $N$ matrices of dimensions $32 \times 32$, using convolutional 2D layers for faster convergence. The inference layers included 3 convolutional 2D layers interweaving with 3 max-pooling layers. The numbers of neurons in the 3 convolutional layers are (32, 64, 128), respectively. The kernel size is 3x3 and the stride is 1. The numbers of neurons in the 4 fully-connected layers were (128, 64, 32, 16), respectively. Finally, a softmax layer was added to produce the inferences for the two dependent variable classes, non-sleepy and sleepy.

The baseline HuBERT-Sleep model was trained on all speech tasks with an 80-20 split for training and testing.

\subsection{HuBERT-Sleep Masking Experiment}
\label{subsec:masking_technique}

In addition to the HuBERT-Sleep baseline model, we used two additional model training techniques. The first training technique was a masking paradigm where certain sets of speech elicitations were eliminated from the training set embeddings. For example, if the Microphone-test task was masked, then the embedding size would be $48 \times 1024$, but one of the 1024 layers would be zeroed out (e.g. $[0,0,\dots, 0]$). In other words, all  \textit{non-microphone-test} task data were used in model training, but the \textit{microphone-test} data were masked. We used this masking technique to train and test models across 12 tasks, as described in Section~\ref{subsec:baseline_model}. By comparing the baseline model--trained with all speech tasks--with a model with a masked speech task, we can evaluate the importance of the masked task on sleepiness inference.

\vspace{-.3cm}
\subsection{HuBERT-Sleep Separate Training Experiment}
\label{subsec:hubert_sleeptask_design}

The second model training technique was to use a small subset of speech task(s) for model training. For example, only the \textit{microphone-test} task was used for model training. For speech tasks that included more than one elicitation, such as the two elicitations in the memory recall task, the embedding size would be $N = 2 \times 1024$. Once again, the separate training models were compared with the baseline model in order to evaluate the importance of speech tasks on sleepiness.

\begin{small}
\begin{table*}[!t]
\label{tbl:sleepy_dist}
\centering\small
\begin{tabular}{|c|c|c|c|}
\hline
\multirow{2}{*}{\textbf{ Task}} & 
                    \multicolumn{3}{c}{Accuracy \% (F1-scores)} \vline\\ \cline{2-4} 
                    & \textbf{\small{TPOT w/ GeMAPs }}
                    & \textbf{\small{Separate w/ HuBERT}} 
                    & \textbf{\small{Masking w/ HuBERT}}\\
                    \cline{1-4}
\centering{T1. Microphone test}& 45.10 (.440) & \textit{\textbf{69.70 (.805)}} & 81.13 (.893) \\
\centering{T2. Free speech}& 54.25 (.578) & 77.24 (.867) & 80.31 (.888)  \\
\centering{T3. Picture description}& 49.67 (.516) & \textit{\textbf{70.66 (.799)}}  & 80.69 (.891) \\
\centering{T4. Category naming}& 42.48 (.333) & 75.00 (.852)  & 81.35 (.895) \\
\centering{T5. Phonemic fluency}& 48.37 (.484) & 78.34 (.876) & 80.02 (.893) \\
\centering{T6. Paragraph reading}& 45.39 (.443) & 73.14 (.840) & 81.62 (.897) \\
\centering{T7. Sustained phonation}& 50.98 (.483) & 77.68 (.870) & 81.56 (.897) \\
\centering{T8. Diadochokinetic (pa-pa-pa)}& 58.17 (.579) & \textit{\textbf{\textbf{67.61 (.822)}}} & 82.00 (.900)  \\
\centering{T9. Diadochokinetic (pa-ta-ka)}& 50.33 (.531) & \textit{\textbf{69.83 (.796)}} & 80.36 (.889) \\
\centering{T10. Confrontational naming}& 44.44-60.13 (.317-.643) & \textbf{81.13 (.894)} & \textbf{78.36 (.889)} \\
\centering{T11. Non-word pronunciation} & 45.10-58.17 (.339-.623) & 78.66 (.877) & 80.85 (.893) \\
\centering{T12. Memory recall}& 46.41-50.98 (.312-.529)& \textbf{80.07 (.853)} & \textbf{72.76 (.886)} \\
\hline
\centering{Baseline (all tasks)}& \textbf{54.90 (.566)} &\multicolumn{2}{c}{\textbf{81.29 (.895)}} \vline\\
\hline
\end{tabular} 
\caption{Accuracy and F1-scores}
\vspace{-0.25cm}
\label{tabl:accuracy_f1}
\end{table*}
\end{small}

\vspace{-0.2cm}

\subsection{The OpenSMILE GeMAPS01a/TPOT Experiment}
\label{subsec:tpot_experiment}
An independent third-party validation was conducted as a way to benchmark the models reported in sections~\ref{subsec:baseline_model} - \ref{subsec:hubert_sleeptask_design}. Instead of using HuBERT embeddings, aggregate acoustic features were used from the OpenSMILE GeMAPSv01a embeddings (62 dimensions per audio file). An automated machine learning tool, TPOT-light, was used to generate the classification models. The TPOT configuration iterates over a range of classification techniques (naive Bayes, decision trees, k-nearest-neighbors, logistic regression) and preprocessors to come up with an optimized model. Forty-eight independent models were created across the 12 tasks. For all models, sleepy and non-sleepy classes were balanced so that the same number of samples were in each class.

\section{Evaluation}
\label{sec:experiment_results}
In this section, the implementation details are presented and the results on sleepiness detection are reported.

\subsection{Implementation details}
\label{subsec:experiment_setup}
The 1,828 sessions from the Voiceome dataset were split into 80\% training (1,462 sessions) and 20\% testing (366 sessions) sets. The training and testing were conducted in 5 non-overlapping rounds by a 40GB of memory Graphics Processing Unit (NVIDIA TESLA A100). The average result of the 5 rounds is reported. In order to adapt our training process to limited memory resources, the learning rate was set to $10^{-4}$, the maximum epoch was set to 200, and the batch size was set to 32.

\subsection{HuBERT-Sleep model}
\label{subsec:experiment_baseline}
The baseline HuBERT-Sleep model was trained and tested on all speech data. The accuracy of this model was 81.29\%, meaning that a person's sleepiness category (sleepy or non-sleepy) was determined correctly 81\% of the time.

The TPOT classification model used MaxAbs scaling for pre-processing the GeMAPS features and a KNN model architecture. The accuracy of this model was 54.90\%, suggesting that the HuBERT-based neural-net architecture was better for detecting sleepiness than other classification methods.

\subsection{Speech task evaluation}
\label{subsec:experiment_results}
Both masking and separate training techniques were used to evaluate the importance of a single speech task. Table~\ref{tabl:accuracy_f1} shows the average accuracy of the test data over 5 rounds; the masking technique and separate training techniques are reported separately. Recall that the masking technique consisted of using all speech data except for the masked task. In Table~\ref{tabl:accuracy_f1}, the masking accuracy score for the \textit{microphone test} indicates the accuracy of the model using all speech data except the \textit{microphone test}. A lower accuracy score for a task indicates that the task was more important for sleepiness detection, as the removal of that task’s data hinders model performance.

On the other hand, the separate training technique uses only the indicated task. The separate training accuracy for the \textit{microphone test} corresponds to the accuracy of the model using only the \textit{microphone test}. For the separate training task, a higher accuracy score is consistent with the idea that the task was more important for sleepiness classification, as that task was the only predictor of sleepiness.

In addition to the masking and separate training methods, we ran non-deep learning tasks with the aggregate GeMAPS features. The first main result is that the HuBERT architecture--both with the masking and separate training techniques--performed with higher accuracy and F1 scores than did the non-deep learning tasks. Of the two HuBERT-based techniques, the masking technique resulted in higher accuracy scores than the separate training technique. 

The accuracy and F1 scores in Table~\ref{tabl:accuracy_f1} are consistent with the idea that the \textit{memory recall} and \textit{category naming} tasks were the most important for predicting sleepiness, as they resulted in the lowest masking scores and highest separate training scores. The results are also consistent with the idea that diadochokinetic tasks (pa-ta-ka; pa-pa-pa), picture descriptions, and the sentence-reading based microphone test are not good predictors of sleepiness, as they produce the lowest separate training scores and high masking scores.

\section{Conclusion}
\label{sec:conclusion}
In this paper, we developed a deep transfer learning model to classify sleepy versus non-sleepy speech samples in the Voiceome dataset. We furthermore utilized two techniques in order to identify the most relevant speech tasks in sleepiness detection. The results of these two training approaches converge: both training techniques suggest that memory recall and categorical naming tasks (from the Boston Naming Test) are the most important tasks for detecting sleepiness from speech data. Future work may wish to employ these two speech tasks in production systems where sleepiness detection can lead to safer and healthier environments. 

\bibliographystyle{IEEEbib}
\bibliography{related, refs}

\end{document}